# GRADED UNIFICATION: A FRAMEWORK FOR INTERACTIVE PROCESSING

Albert Kim [*]
Department of Computer and Information Sciences
University of Pennsylvania
Philadelphia, Pennsylvania, USA
email: alkim@unagi.cis.upenn.edu

## Abstract

An extension to classical unification, called *graded unification* is presented. It is capable of combining contradictory information. An interactive processing paradigm and parser based on this new operator are also presented.

## Introduction

Improved understanding of the nature of knowledge used in human language processing suggests the feasibility of interactive models in computational linguistics (CL). Recent psycholinguistic work such as (Stowe, 1989; Trueswell et al., 1994) has documented rapid employment of semantic information to guide human syntactic processing. In addition, corpus-based stochastic modelling of lexical patterns (see Weischedel et al., 1993) may provide information about word sense frequency of the kind advocated since (Ford et al., 1982). Incremental employment of such knowledge to resolve syntactic ambiguity is a natural step towards improved cognitive accuracy and efficiency in CL models.

This exercise will, however, pose difficulties for the classical ('hard') constraint-based paradigm. As illustrated by the Trueswell et al. (1994) results, this view of constraints is too rigid to handle the kinds of effects at hand. These experiments used pairs of locally ambiguous reduced relative clauses such as:

*1) the man recognized by the spy took off down the street*
*2) the van recognized by the spy took off down the street*

The verb *recognized* is ambiguously either a past participial form or a past tense form. Eye tracking showed that subjects resolved the ambiguity rapidly (before reading the *by*-phrase) in *2)* but not in *1)* [1]. The conclusion they draw is that subjects use knowledge about thematic roles to guide syntactic decisions. Since *van*, which is inanimate, makes a good Theme but a poor Agent for *recognized*, the past participial analysis in *2)* is reinforced and the main clause (past tense) suppressed. Being animate, *man* performs either thematic role well, allowing the main clause reading to remain plausible until the disambiguating *by*-phrase is encountered. At this point, readers of *1)* displayed confusion.

Semantic constraints do appear to be at work here. However, the effects observed by Trueswell et al. are graded. Verb-complement combinations occupy a continuous spectrum of "thematic fit", which influences reading times. This likely stems from the variance of verbs with respect to the thematic roles they allow (e.g., Agent, Instrument, Patient, etc.) and the syntactic positions of these.

The upshot of such observations is that classical unification (see Shieber, 1986), which has served well as the combinatory mechanism in classical constraint-based parsers, is too brittle to withstand this onslaught of uncertainty.

This paper presents an extension to classical unification, called *graded unification*. Graded unification combines two feature structures, and returns a *strength* which reflects the compatibility of the information encoded by the two structures. Thus, two structures which could not unify via classical unification may unify via graded unification, and all combinatory decisions made during processing are endowed with a level of goodness. The operator is similar in spirit to the operators of fuzzy logic (see Kapcprzyk, 1992), which attempts to provide a calculus for reasoning in uncertain domains. Another related approach is the "Unification Space" model of Kempen & Vosse (1989), which unifies through a process of simulated annealing, and also uses a notion of unification strength.

A parser has been implemented which combines constituents via graded unification and whose decisions are influenced by unification strengths. The result is a paradigm of incremental processing, which maintains a feature-based system of knowledge representation.

## System Description

Though the employment of graded unification engenders a new processing style, the system's architecture parallels that of a conventional unification-based parser.

### Feature Structures: Prioritized Features

The feature structures which encode the grammar in this system are conventional feature structures augmented by the association of *priorities* with each atomic-valued feature. Prioritizing features allows them to vary in terms of influence over the strength of unification. The priority of an atomic-valued feature $f_i$ in a feature structure $X$ will be denoted by $Pri(f_i, X)$. The effect of feature prioritization is clarified in the following sections.

---

[*] I thank Christy Doran, Jason Eisner, Jeff Reynar, and John Trueswell for valuable comments. I am grateful to Ewan Klein and the Centre for Cognitive Science, Edinburgh, where most of this work was conducted, and also acknowledge the support of DARPA grant N00014-90-J-1863.

[1] In fact, ambiguity effects were often completely eliminated in examples like *2)*, with reading times matching those for the unambiguous case:
*3) the man/van that was recognized by the spy ...*

## Graded Unification

Given two feature structures, the graded unification mechanism ($\sqcup_G$) computes two results, a unifying structure and a unification strength.

**Structural Unification** Graded unification builds structure exactly as classical unification except in the case of atomic unification, where it deviates crucially.

Atoms in this framework are weighted disjunctive values. The weight associated with a disjunct is viewed as the confidence with which the processor believes that disjunct to be the 'correct' value. Figures 1(a) and 1(b) depict atoms (where 1(a) is "truly atomic" because it contains only one disjunct).

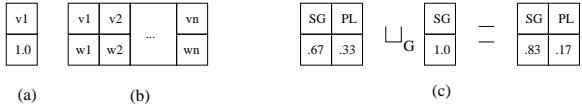

Figure 1: Examples of Atoms

Atomic unification creates a mixture of its two argument atoms as follows. When two atoms are unified, the set union of their disjuncts is collected in the result. For each disjunct in the result, the associated weight becomes the average of the weights associated with that disjunct in the two argument atoms. Figure 1(c) shows an example unification of two atoms. The result is an atom which is 'believed' to be SG (singular), but could possibly be PL (plural).

**Unification Strength** The unification strength (denoted $\sqcup_G Strength$) is a weighted average of atomic unification strengths, defined in terms of two sums, the *actual compatibility* and the *perfect compatibility*.

If $A$ and $B$ are non-atomic feature structures to be unified, then the following holds:

$$\sqcup_G Strength(A,B) = \frac{ActualCompatibility(A,B)}{PerfectCompatibility(A,B)}.$$

The **actual compatibility** is the sum:

$$\sum_i \begin{cases} \frac{Pri(f_i,A)+Pri(f_i,B)}{2} * \sqcup_G Strength(v_{iA}, v_{iB}) \\ \quad \text{if } f_i \text{ shared by } A \text{ and } B \\ Pri(f_i, A) \quad \text{if } f_i \text{ occurs only in } A \\ Pri(f_i, B) \quad \text{if } f_i \text{ occurs only in } B \end{cases}$$

where $i$ indexes all atomic-valued features in $A$ or $B$, and $v_{iA}$ and $v_{iB}$ are the values of $f_i$ in $A$ and $B$ respectively. The **perfect compatibility** is computed by a formula identical to this except that $\sqcup_G Strength$ is set to 1.

If $A$ and $B$ are atomic, then $\sqcup_G Strength(A,B)$ is the total weight of disjuncts shared by $A$ and $B$: $\sqcup_G Strength(A,B) = \sum_i Min(w_{iA}, w_{iB})$ where $i$ indexes all disjuncts $d_i$ shared by $A$ and $B$, and $w_{iA}$ and $w_{iB}$ are the weights of $d_i$ in $A$ and $B$ respectively.

By taking atomic unification strengths into account, the actual compatibility provides a raw measure of the extent to which two feature structures agree. By ignoring unification strengths (assuming a value of 1.0), the perfect compatibility is an idealization of the actual compatibility; it is what the actual compatibility would be if the two structures were able to unify via classical unification. Thus, unification strength is always a value between 0 and 1.

## The Parser: Activated Chart Edges

The parser is a modified unification-based chart parser. Chart edges are assigned *activation levels*, which represent the 'goodness' of (or confidence in) their associated analyses. Each new edge is activated according to the strength of the unification which licenses its creation and the activations of its constituent edges.

**Constraining Graded Unification** Without some strict limit on its operation, graded unification will overgenerate wildly. Two mechanisms exist to constrain graded unification. First, if a particular unification completes with strength below a specified *unification threshold*, it fails. Second, if a new edge is constructed with activation below a specified *activation threshold*, it is not allowed to enter the chart, and is *suspended*.

**Parsing Strategy** The chart is initialized to contain one inactive edge for each lexical entry of each word in the input. Lexical edges are currently assigned an initial activation of 1.0.

The chart can then be expanded in two ways:

1. An active edge may be extended by unifying its first unseen constituent with the LHS of an inactive edge.

2. A new active edge may be created by unifying the LHS of a rule with the first unseen constituent of some active edge in the chart (top down rule invocation).

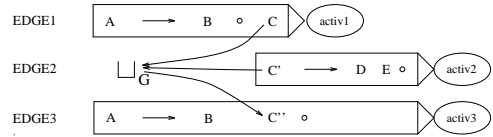

Figure 2: Extension of an Active Edge by an Inactive Edge

Figure 2 depicts the extension of the active EDGE1 with the inactive EDGE2. The characters represent feature structures, and the ovular nodes on the right end of each edge represent activation level. The parser tries to unify $C'$, the mother node of EDGE2, with $C$, the first needed constituent of EDGE1. If this unification succeeds, the parser builds the extended edge, EDGE3 (where $C \sqcup_G C'$ produces $C''$). The activation of the new edge is a function of the strength of the unification and the current activations of EDGE1 and EDGE2:

$$activ_3 = w_1 \cdot \sqcup_G STRENGTH(C,C')$$
$$+ w_2 \cdot activ_1$$
$$+ w_3 \cdot activ_2 \quad \text{(The weights } w_i \text{ sum to 1.)}$$

EDGE3 enters the chart only if its activation exceeds the activation threshold. Rule invocation is depicted in figure 3. The first needed constituent in EDGE1 is unified with the LHS of RULE1. EDGE2 is created to begin searching for $C$. The new edge's activation is again a function of unification strength and other activations:

$$activ_3 = w_1 \cdot \sqcup_G STRENGTH(C,C')$$
$$+ w_2 \cdot activ_1$$
$$+ w_3 \cdot activ_2$$

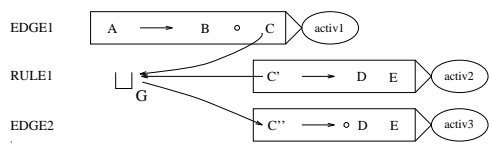

Figure 3: Top Down Rule Invocation

The activation levels of grammar rule edges, like those for lexical edges, are currently pegged to 1.0.

## A Framework for Interactive Processing

The system described above provides a flexible framework for the interactive use of non-syntactic knowledge.

### Animacy and Thematic Roles

Knowledge about animacy and its important function in the filling of thematic roles can be modelled as a binary feature, ANIMATE. A (active voice) verb can strongly 'want' an animate Agent by specifying that its subject be [ANIMATE +] and assigning a high priority to the feature ANIMATE. Thus, any parse combining this verb with an inanimate subject will suffer in terms of unification strength. A noun can be strongly animate by having a high weight associated with the positive value of ANIMATE. Animacy has been encoded in a toy grammar. However, principled settings for the priority of this feature are left to future work.

### Statistical Information from Corpora

Corpus-based part-of-speech (POS) statistics can also be naturally incorporated into the current model. It is proposed here that a Viterbi decoder could be used to generate the likelihoods of the $n$ best POS tags for a given word in the input string. Lexical chart edges would then be initially activated to levels proportional to the predicted likelihoods of their associated tags. Since these activations will be propagated to larger edges, parses involving predicted word senses would consequently be given a head start in a race of activations. Attractively, this strategy allows a fuller use of statistical information than one which uses the information simply to deterministically choose the $n$ best tags, which are then treated as equally likely.

### Interaction of Diverse Information

A crucial feature of this framework is its potential for modelling the interaction between sources of information like the two above when they disagree. Sentences 1) and 2) again provide illustration. In such sentences, knowledge about word sense frequency supports the wrong analysis, and semantic constraints must be employed to achieve the correct (human) performance.

Intuitively, the raw frequency (without considering context) of the past tense form of *recognized* is higher than that of the past participial. POS taggers, despite considering local context, consistently mis-tag the verb in reduced relatives. The absence of a disambiguating relativizer (e.g., *that*) is one obvious source of difficulty here. But even the ostensibly disambiguating preposition *by*, is itself ambiguous, since it might introduce a manner or locative phrase consistent with the main clause analysis. [2]

Modelling human performance in such contexts requires allowing thematic information to compete against and defeat word frequency information. The current model allows such competition, as follows. POS information may incorrectly predict the main clause analysis, boosting the lexical edge associated with the past tense, and thereby boosting the main clause parse. However, the unification combining the past tense form of *recognized* with an inanimate subject (*van*) will be weak, due to the constraints encoded in the verb's lexical entry. Since the activations of constituent edges depend on the strengths of the unifications used to build them, the main clause parse will lose activation. The parse combining the past participial with an inanimate subject (Theme) will suffer no losses, allowing it to overtake the incorrect parse.

## Conclusions and Future Work

Assigning feature priorities and activation thresholds in this model will certainly be a considerable task. It is hoped that principled and automated methods can be found for assigning values to these variables. One promising idea is to glean information about patterns of subcategorization and thematic roles from annotated corpora. Annotation of such information has been suggested as a future direction for the Treebank project (Marcus *et al.*, 1993). It should be noted that learning such information will require more training data (hence larger corpora) than learning to tag part of speech. In addition, psycholinguistic studies such as the large norming study [3] of MacDonald and Pearlmutter (described in Trueswell *et al.*, 1994) may prove useful in encoding thematic information in small lexicons.

---

[2] In fact, the utility of *by* is neutralized in the case of POS tagging, since prepositions are uniformly tagged (e.g., using the tag IN in the Penn Treebank; see Marcus *et al.*, 1993).

[3] These studies attempt to establish thematic patterns by asking large numbers of subjects to answer questions like "How typical is it for a van to be recognized by someone?" with a rating between 1 and 7.